# Computationally Efficient Data-Driven MPC for Agile Quadrotor Flight

Wonoo Choo and Erkan Kayacan

*Abstract*— This paper develops computationally efficient data-driven model predictive control (MPC) for Agile quadrotor flight. Agile quadrotors in high-speed flights can experience high levels of aerodynamic effects. Modeling these turbulent aerodynamic effects is a cumbersome task and the resulting model may be overly complex and computationally infeasible. Combining Gaussian Process (GP) regression models with a simple dynamic model of the system has demonstrated significant improvements in control performance. However, direct integration of the GP models to the MPC pipeline poses a significant computational burden to the optimization process. Therefore, we present an approach to separate the GP models to the MPC pipeline by computing the model corrections using reference trajectory and the current state measurements prior to the online MPC optimization. This method has been validated in the Gazebo simulation environment and has demonstrated of up to $50\%$ reduction in trajectory tracking error, matching the performance of the direct GP integration method with improved computational efficiency.

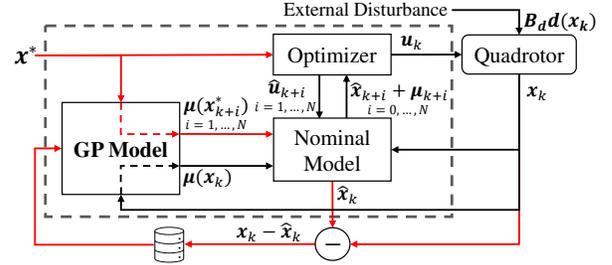

Fig. 1: Architecture of our computationally efficient GP-based MPC. Off-line computations are denoted in red and online computations are denoted in black.

## I. INTRODUCTION

Many modern control methods rely on an accurate dynamic model of the plant. In particular, Model Predictive Control (MPC) is a popular model-based control method that is commonly used for trajectory control for unmanned systems, such as mobile robots, drones, robotic boats, and self-driving cars [1–4]. MPC is an optimization-based technique that utilizes the system model to predict the system behavior to generate the optimal trajectory into the future of a finite horizon. This allows the controller to enforce constraints and avoid collisions while setting system limitations at each time step(s) [2, 5].

As these control methods utilize system models to predict the behavior of systems, the accuracy of the model heavily determines the performance of the control method [6]. Overly complex models allow for more accurate predictions of the system. However, it can render the model itself a computationally incalculable problem. In juxtaposition, an overly simplified system model can result in poor tracking performance and even instability. Therefore, for real-time applications, the dynamic model of the system must be developed to ensure an intersection of balance is met between accuracy and computational feasibility. This intersection is easily met for simpler environments with lower-order mathematical systems. However, systems operating with a greater number of environmental variables can result in incalculable higher-order dynamics [4, 7]. These unmodelled dynamics can lead to inaccurate predictions of the system's response, resulting in poor trajectory tracking performance [8]. Agile quadrotors experience a high level of disturbances such as turbulent effects from the fuselage, the interaction between propellers, and downwash of other propellers [4]. All of which are difficult to model due to their complexities [4].

Disregarding the high-order effects and treating them as external disturbances allows for an efficient MPC; however, during speeds above $5ms^{-1}$ and accelerations above $2G$, its tracking performance is significantly compromised [4]. There are lines of research that investigate deep neural networks to learn the dynamics of the entire system for a number of applications such as cars, fluids, or robot arms to supplant the nominal dynamics in the MPC [4, 9–11]. Though the resulting dynamics are expressive, the high level of nonlinear nature of these models can result in the presence of local minima, which can cause optimization problems to become intractable and hinder the accuracy of control decisions [4]. This can be overcome by using a sampling-based optimizer; however, the computational feasibility scales poorly with the increase in the dimensionality of the input space [4]. Therefore, instead of learning the full dynamics from the data, the nominal model is combined with Gaussian Process (GP) regression models [4, 5, 7]. The residual dynamics that are challenging to model and are neglected are learned by the GPs. This significantly simplifies the learning problem and allows the dimensionality of the machine learning model to be limited.

Authors of [4] incorporate the GP regression models in their MPC optimization pipeline. Therefore, the system's behavior in the control horizon is predicted using the GP augmented dynamics. This accurate prediction comes at the cost of computing GP predictions at every iteration of the MPC optimization, which can be a huge computational burden in real-time applications.

In this work, we propose utilizing the reference trajectory to compute the GP predictions prior to the MPC optimization

Wonoo Choo and Erkan Kayacan are with the School of Aerospace and Mechanical Engineering, The University of Oklahoma, Norman, OK 73019, USA. Email: {w.choo, erkan}@ou.edu

to reduce the computational complexity without compromising the tracking performance of the controller. Only the first node in the control horizon is corrected on-line by the GPs using the current state measurements. The remaining nodes are corrected using the predictions computed offline prior to solving the online MPC. This leverages the controller's capability to track the reference trajectory by assuming that the reference trajectory is a suitable approximation of the system's states in the control horizon. It is evident in the simulation results that this method can reduce the computational cost on-line while achieving a similar level of improvements in tracking performance when compared to [4].

The paper is organized as follows: The system model for a quadrotor, and the traditional formulations of MPC and GP are given in Section II. The developed methodology for data collection, model learning and the formulations of GP augmented MPC is given in Section III. The simulation studies are given in Section IV. Then, a brief conclusion is finally drawn in Section V.

## II. PRELIMINARIES

### A. Notation

In this paper, the scalars are denoted with a lowercase, vectors with a lowercase bold, and matrices with an uppercase bold. The Euclidean vector norm is denoted as $\|\cdot\|$. The axis system of the World and Body frames are as shown in Fig. 2. The Body frame is positioned at the center of mass of the quadrotor with the assumption that the rotors are on the $xy$-plane of the Body frame. A vector from coordinate $p_1$ to $p_2$ expressed in the World frame is denoted as $_W v_{12}$. If $p_1$ is the origin of the frame it is described in, then the frame index is omitted. The attitude state of the quadrotor is represented using unit quaternion, $q_{WB} = (q_w, q_x, q_y, q_z)$ where $\|q_{WB}\| = 1$. The frame rotation by quaternion is given by the quaternion-vector product, $\odot$, such that $q \odot v = qv\bar{q}$, where $\bar{q}$ is the conjugate of $q$.

### B. System Dynamic Model

The quadrotor is modeled as a 6 degree-of-freedom rigid body with diagonal moment of inertia matrix $J = diag(J_x, J_y, J_z)$ and the control inputs $u$ as the rotor thrusts $T_i \, \forall i \in [0, 3]$. The quadrotor's position in the World frame is given by $p_{WB}$, its attitude state of the body is given by $q_{WB}$, its linear velocities in the World frame is denoted by $v_{WB}$ and its angular velocity is given by $\omega_B$. The thrusts on each rotor are modeled individually such that the collective thrust, $T_B$, and the body torque, $\tau_B$, on the quadrotor, is given by

$$T_B = \begin{bmatrix} 0 \\ 0 \\ \sum T_i \end{bmatrix} \text{ and } \tau_B = \begin{bmatrix} d_y(-T_0 - T_1 + T_2 + T_3) \\ d_x(-T_0 + T_1 + T_2 - T_3) \\ c_\tau(-T_0 + T_1 - T_2 + T_3) \end{bmatrix} \quad (1)$$

where $d_x$ and $d_y$ are the rotor displacements from the center of the mass of the quadrotor and $c_\tau$ is the rotor drag torque constant. The nonlinear state-space model of the system is 13-dimensional, and its dynamics are given by the following

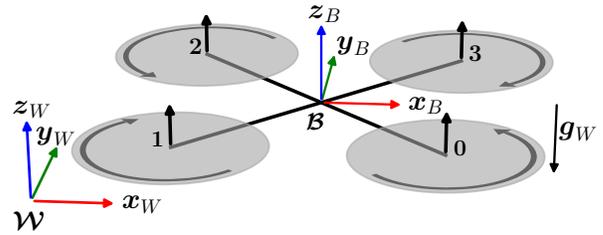

Fig. 2: Visualization of the quadrotor with the World and Body frames.

equation

$$\dot{x} = f_{dyn}(x, u) = \begin{bmatrix} v_W \\ q_{WB} \cdot \begin{bmatrix} 0 \\ \omega_B/2 \end{bmatrix} \\ q_{WB} \odot T_B + g_W \\ J^{-1}(\tau_B - \omega_B \times J\omega_B) \end{bmatrix} \quad (2)$$

where $x = [p_{WB}, q_{WB}, v_{WB}, \omega_B]^\intercal$ is the state vector and $g_W$ denotes the Earth's gravity. These dynamics were realized in discrete-time steps $\delta t$ given an initial state $x_k$ and input $u_k$ utilizing explicit $4^{th}$ order Runge-Kutta method where the discretized dynamics are denoted as $f_{RK4}$ as shown in (3). Additionally, we consider a dynamic model of the system that is composed of known nominal dynamics, $f_{RK4}$, and unknown dynamics neglected by the nominal term, $d$. These unknown dynamics are assumed to be in the subspace spanned by $B_d$ and will be learned by the GP models.

$$x_{k+1} = f_{RK4}(x_k, u_k, \delta t) + B_d d(x_k) \quad (3)$$

### C. MPC Formulation

MPC is an optimization-based control method that determines an optimal plan of action over a fixed time horizon [12]. The prediction horizon, $T$, is discretized into $N$ control nodes to compute an optimal trajectory obeying the constraints defined by the user. Its primary applications include tracking and stabilization of a system under its dynamics $x_{k+1} = f_{RK4}(x_k, u_k, \delta t)$ along a reference trajectory $x^*(t), u^*(t)$ by minimizing an objective function, also known as the cost function.

The objective function can take multiple forms allowing the user to shape the response of the MPC to suit the requirements of the applications. Here we define the cost function to be of the quadratic form to minimize the error between the system states and the reference trajectory as shown in (4). Here no obstacles were simulated; thus, only the input limitations were accounted for.

$$\min_{u} \|x_N - x_N^*\|_{Q_T}^2 + \sum_{k=0}^{N} \|x_k - x_k^*\|_Q^2 + \|u_k - u_k^*\|_R^2 \quad (4a)$$

$$\text{s.t.} \quad x_{k+1} = f_{RK4}(x_k, u_k, \delta t) \quad (4b)$$

$$x_0 = x_{init} \quad (4c)$$

$$u_{min} \leq u_k \leq u_{max} \quad (4d)$$

This was formulated using ACADOS [13], and CasADi [14] as a multiple shooting problem solving a sequential quadratic program in a real-time iterating scheme.

## D. Gaussian Process Regression

Like the majority of supervised machine learning algorithms, GP tries to fit an unknown function to a set of training data [15]. Here we utilise GP to identify the unknown dynamics of the system $d: \mathbb{R}^{n_z} \to \mathbb{R}^{n_d}$ from a set of inputs $z_k \in \mathbb{R}^{n_z}$ and outputs $y_k \in \mathbb{R}^{n_d}$:

$$y_k = d(z_k) + w_k \tag{5}$$

where the process noise $w_k \sim \mathcal{N}(\mathbf{0}, \mathbf{\Sigma})$ is independent and identically distributed Gaussian noise with diagonal covariance $\mathbf{\Sigma} = \text{diag}([\sigma_1^2, ..., \sigma_{n_d}^2])$. This allows each dimension of $y_k$ to be treated independently through individual 1-dimensional GPs. Given the training samples $z$ and the test point $z_*$, the mean and variance of the GP prediction are given by:

$$\mu(z_*) = \mathbf{k}_*^\intercal \mathbf{K}^{-1} \mathbf{y} \quad \Sigma_{\mu k} = k_{**} - \mathbf{k}_*^\intercal \mathbf{K}^{-1} \mathbf{k}_* \tag{6}$$
$$\text{with} \quad \mathbf{K} = \kappa(\mathbf{z}, \mathbf{z}) + \sigma_n^2 \mathbf{I}$$
$$\mathbf{k}_* = \kappa(\mathbf{z}, z_*) \quad k_{**} = \kappa(z_*, z_*)$$

where $\mathbf{K}$ denotes the covariance between the training samples, also known as the Gram matrix. $\mathbf{k}_*$ denotes the covariance between the training sample and the test point and $k_{**}$ denotes the variance of the test point. Here we utilize the Radial Basis Function (RBF) kernel defined by (7) to compute these (co)-variance.

$$\kappa(z_i, z_j) = \sigma_f^2 \exp\left(-\frac{1}{2}(z_i - z_j)^\intercal \mathbf{L}^{-2}(z_i - z_j)\right) + \sigma_n^2 \tag{7}$$

where $z_i$ and $z_j$ represent the data features, $\mathbf{L}$ is the diagonal length scale matrix, and $\sigma_f$ and $\sigma_n$ represent the data and prior noise variance, respectively [4, 16]. The variables $\mathbf{L}$, $\sigma_f$, and $\sigma_n$ shape the response of the GP regression model and are optimized during model training. The computational complexity of the GP prediction is given by $\mathcal{O}(n^3)$ where $n$ is the number of data points. This motivates the use of sparse approximation methods using inducing points to minimize the training sample size.

There are many approximation methods that reduce the computational complexity of the direct implementation of the GP. We implement the sparse GP regression technique introduced in [17], where the algorithm is reinterpreted with approximated priors rather than approximating the inferences [17]. This approach analyzes the posterior and then computes the *effective prior* that is then used to encapsulate the behaviour of the original GP into a smaller model. The computational complexity of the sparse GP is given by $\mathcal{O}(m^3)$ where $m$ is the number of effective prior and $m \ll n$ [17].

## III. DATA-DRIVEN MPC

### A. Data Collection and Model Learning

The training samples for the GP models are collected by running the nominal MPC on the given lemniscate trajectory. The velocity of the quadrotor is ramped up to an axial velocity of $8ms^{-1}$ and back down to rest. At each sample time $t_k$, the time step $\delta t_k$, the next sample time's actual velocity $_B v_{k+1}$, and the predicted velocity $_B \hat{v}_{k+1}$ are recorded. Then, the acceleration error that the nominal model does not capture is computed by:

$$_B a_{ek} = \frac{_B v_{k+1} - _B \hat{v}_{k+1}}{\delta t_k} \tag{8}$$

The training samples were all transformed into the body frame. The GP models were trained such that the body frame velocity $_B v$ was mapped to the body frame acceleration disturbances $_B a_e$. The residual dynamics in each axis were modeled independently, resulting in three separate GP models for $v_x$, $v_y$, and $v_z$. This was done to minimize the required inducing points. The GP predictions can be written as:

$$_B a_{ek} = \boldsymbol{\mu}_v(_B v_k) = \begin{bmatrix} \mu_{vx}(_B v_{xk}) \\ \mu_{vy}(_B v_{yk}) \\ \mu_{vz}(_B v_{zk}) \end{bmatrix}$$
$$\boldsymbol{\Sigma}_\mu(_B v_k) = \text{diag}\left(\begin{bmatrix} \Sigma_{vx}(_B v_{xk}) \\ \Sigma_{vy}(_B v_{yk}) \\ \Sigma_{vz}(_B v_{zk}) \end{bmatrix}\right) \tag{9}$$

### B. Gaussian Process Augmented Control Formulation

As done in [4], we utilize the GP models to correct the nominal model of the quadrotor. But rather than integrating the GP model with the nominal model in the MPC pipeline, we separate the GP predictions with the MPC optimization to minimize the computational burden. The MPC is formulated to take in model correction parameters, $\hat{d}$, that are updated at each time-steps. The GPs correct the model dynamics in the first control node online using the current state measurements. The corrections for the remaining nodes are computed prior to the flight using the reference trajectory. Only the corrections to the first node are computed online as it is the most critical node in the MPC. The first control action in the MPC is applied to the system where the remaining control elements are disregarded, highlighting the importance of the first node. Furthermore, as the predicted states will converge to the reference trajectory in the control horizon, the differences between the predicted states and the reference trajectory will decrease. The augmented MPC formulation is given by:

$$\min_{\boldsymbol{u}} \|\boldsymbol{x}_N - \boldsymbol{x}_N^*\|_{\boldsymbol{Q}_T}^2 + \sum_{k=0}^{N} \|\boldsymbol{x}_k - \boldsymbol{x}_k^*\|_{\boldsymbol{Q}}^2 + \|\boldsymbol{u}_k - \boldsymbol{u}_k^*\|_{\boldsymbol{R}}^2$$
$$\text{s.t.} \quad \boldsymbol{x}_{k+1} = \boldsymbol{f}_{RK4}(\boldsymbol{x}_k, \boldsymbol{u}_k, \delta_t) + \boldsymbol{B}_d \hat{\boldsymbol{d}}_k$$
$$\boldsymbol{x}_0 = \boldsymbol{x}_{init}$$
$$\hat{\boldsymbol{d}}_0 = \boldsymbol{\mu}(\boldsymbol{B}_z \boldsymbol{x}_0)$$
$$\hat{\boldsymbol{d}}_k = \boldsymbol{\mu}(\boldsymbol{B}_z \boldsymbol{x}_k^*) \ \forall \ k = 1, ..., N-1$$
$$\boldsymbol{u}_{min} \leq \boldsymbol{u}_i \leq \boldsymbol{u}_{max}$$
$$\tag{10}$$

where the input states to the GPs are selected by $\boldsymbol{B}_z$ and the corrected states are selected by $\boldsymbol{B}_d$.

## IV. SIMULATION STUDIES

### A. Simulation Setup

The performance of the controllers has been tested and validated in a Robot Operating System (ROS) Gazebo [18] environment. The AscTec Hummingbird quadrotor model has been simulated using the RotorS extension [19]. The aerodynamic effect simulated by the RotorS package is the rotor drag which exhibits a linear relationship to the body frame velocity [19].

The experiment is conducted with quadrotors following a lemniscate trajectory of $[x(t) = 5\cos(\sqrt{2}t) - 5, y(t) = 5\sin(\sqrt{2}t)\cos(\sqrt{2}t), z(t) = 2.5]$ as shown in Fig. 3. The reference trajectory accelerates the quadrotor from the hovering state, reaching a maximum velocity of $11.24 ms^{-1}$ and then decelerates back down to a hovering position.

The MPC is implemented with a prediction horizon of 1 second. The method developed by [4] performs GP predictions for each node online, which can be computationally expensive with a larger number of nodes in the prediction horizon. Therefore, the number of nodes is set to 10 to reduce the computational cost for real-time implementation.

The MPC is computed at a frequency of $50 Hz$, and its collective thrust and desired body rates are then taken as input to the RPG controller [20, 21] which controls the rotor thrusts. All system states were measured at a frequency of $100 Hz$ and cascaded with noise. The added noise on the system signals is zero-mean Gaussian distributed noise with a standard deviation of $7mm$, $0.4°$, $7mms^{-1}$ and $0.4°s^{-1}$ of the state measurements.

We studied two implementations of GP augmented MPC: *GPy-MPC* (developed in this paper) and *GP-MPC* from [4]. Both GP models in each implementation were trained from the same dataset that were collected as described in Section III-A. The two implementations only differ in the integration of the GP models to the MPC framework and the packages utilized for GP models. The GP-MPC incorporates the GP augmented dynamic model of the system into the MPC pipeline to optimize the control trajectory with the GP corrections. To achieve this the authors of [4] developed a direct implementation of GP. Instead, we separated the GP predictions from the optimization of the MPC to minimize the computational burden of the GP. We achieved this by approximating the predicted states in the control horizon with the reference trajectory to pre-compute the corrections from the GPs. We utilized the open-source package GPyTorch [22] to implement our GP models. The GPyTorch package is optimized for training and predicting large data-sets rather than making numerous small predictions [22]. As the GP must compute the model correction for the entire trajectory, GPyTorch package was chosen over the direct implementation from [4]. Moreover, GPyTorch package produced GP models that better regressed the collected data-set. These two augmented MPCs were compared to an MPC with just the *Nominal* model to analyze their improvements in closed-loop position tracking results.

The simulation studies were conducted on a laptop with 16GB of RAM, 10th Generation Intel Core i7-10750H, and NVIDIA GeForce RTX 2070 8GB GDDR6.

### B. Simulation Results of Gaussian Process Regression Models

The trained GP regression models mapping the aerodynamic effects on the quadrotor are visualized in Fig. 4. The models shown in Fig. 4 are sparse GP models trained by utilizing 20 inducing points derived from dense GP models, that were trained from 400 sub-sampled data points. The model trained using the GPyTorch package is denoted in orange and the model trained using the source code from [4] is denoted in black.

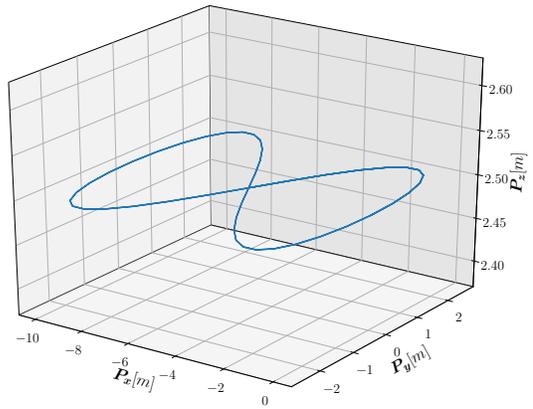

Fig. 3: Trajectory of the quadrotor

The complete data-set consisting of 10200 points were organized into 400 bins categorized by their initial velocity. Then the median values were selected from each bin to form the training points for the dense GPs. To select the inducing points for the sparse GPs; the original data-set were organized into 20 bins, and the mean velocity of each bin was selected to be the inducing locations. The acceleration errors of these inducing locations were then predicted using the dense GPs to compute the effective prior. This allowed the dense GPs to capture the overall behavior of the unmodelled dynamics and the sparse GPs to imitate the regression of the dense GPs. The sparse GP was chosen to be trained using 20 inducing points as the optimal range was found to be in the range of 15 to 25 samples in regards to the computational complexity and the control performance [4].

It is evident that for a noise-free simulation, both GP models were able to closely map the relationship between the body-frame velocity to body-frame acceleration error. Despite the significant variance present in the noisy data, both GP models were able to map the general trend of the data in $x$ and $y$ axes. Moreover, the response of the GPs of the noisy simulation does not differ greatly from the noise-free simulation. This demonstrates the robustness of the GPs to Gaussian distributed noise.

The responses of both GP models follow a similar trend for the $x$ and $y$ axes. However, the data in the $z$ axis do not present a strong trend. The two GPs in the $z$ axis exhibit a similar behavior at lower velocities and diverge at higher velocities. These differences may be the source of the slight differences in the performance of the two controllers.

### C. Simulation Results of GP Augmented MPC

The Root Mean Squared Error (RMSE) of the tracking performance of the three controllers are tabulated in Table I along with the average optimization time of each algorithm. The closed-loop tracking error of the controllers in a noisy simulation as a function of maximum velocity achieved is shown in Fig. 5.

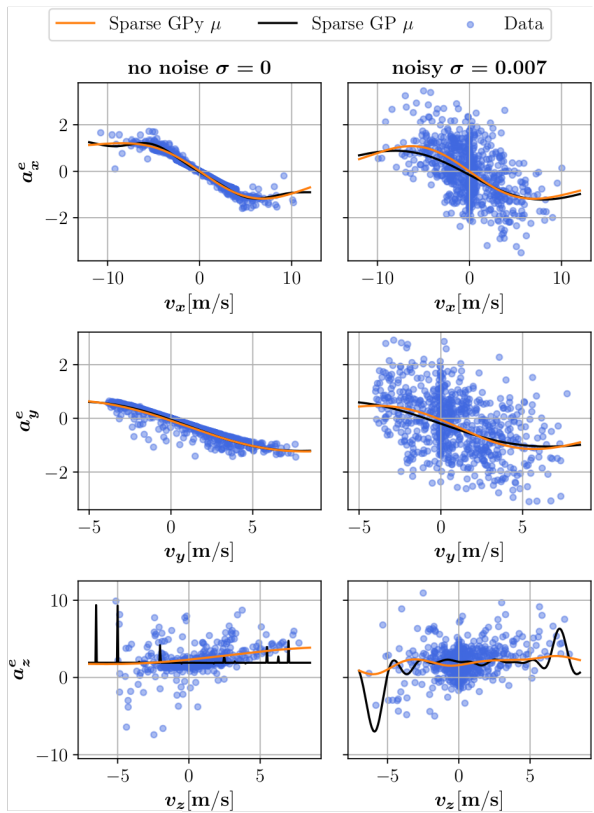

Fig. 4: GP models mapping the body frame velocity to body frame acceleration error in simulation environment with varying noise levels. The GP models trained using the package from the literature are denoted in black and the GP models trained using GPyTorch package are denoted in orange.

Due to the aggressiveness of the trajectory and the rapidly changing velocities, nominal MPC struggled to accurately control the quadrotor with limited knowledge of the system. It is to be noted that the given simulation environment only simulates the rotor drag, which exhibits a linear relationship with the body frame velocity. Therefore, in a real-time application where the unmodelled disturbances exceed the limited simulation capabilities of the conducted test, the inaccuracy of the nominal MPC will compound. GP-MPC and GPy-MPC showed improvements of up to 65% and 50% in a noise-free simulation, respectively. It was expected that the GP-MPC display a superior improvement over GPy-MPC as their implementation provides a more accurate input to the GPs. However, the difference of $2cm$ in tracking errors in the real-time application could be negligible. In a real-time application, the sensor noise and its resolution may mask the $2cm$ difference in the control performance. Moreover, when the RMSE of the tracking error was calculated from only $p_x$ and $p_y$, both controllers returned a RMSE of $39mm$. This shows that the difference in the performance of the controllers is from the correction of the GP in the $z$ direction where the two GP models did not exhibit a similar trend compared to the other axes.

In a noisy simulation environment GP-MPC and GPy-

TABLE I: Comparison of closed-loop trajectory tracking results and the average optimization time of GP-MPC and GPy-MPC methods in simulation environment with varying noise levels reaching velocities of up to $11.24 ms^{-1}$

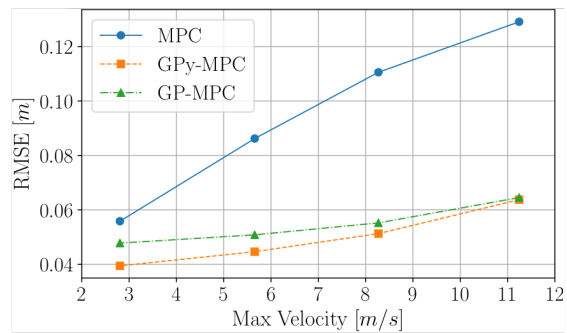

|  | Model | | | | |
|---|---|---|---|---|---|
|  | Nominal | GP-MPC | | GPy-MPC | |
| Noise Level $\sigma$ | RMSE [mm] | RMSE [mm] | %↓ | RMSE [mm] | %↓ |
| 0 | 127.7 | 44.1 | 65 | 63.9 | 50 |
| 0.007 | 129.1 | 64.5 | 50 | 63.7 | 50 |
| **Opt. Time** [ms] | 1.489 | 4.527 | | 3.452 | |

Fig. 5: Comparison of tracking error of the controllers at different max velocities achieved with noise present.

MPC both showed improvements of up to 50% reduction in their tracking error. As hypothesized the tracking error between GP-MPC and GPy-MPC in a noisy environment is negligible. Despite the significant reduction in the number of operations of the GPs, the average optimization time has not decreased notably. The average optimization time of the nominal MPC, GP-MPC , and GPy-MPC are given by 1.489 ms, 4.527 ms and 3.452 ms respectively. The computation time of the GPy-MPC has been improved by 23% (1ms) with a prediction horizon of 10 nodes. This is due to the GPyTorch package being optimized for larger data-set rather than single-point predictions. The advantages of using the GPyTorch package are that it is able to produce improved regression than the package from [4] and is optimal for large offline computation. However, it is sub-optimal for computing single-point predictions that are made online. Regardless, this shows that utilizing the reference trajectories can be a valid approach to improving the computational efficiencies of the GP framework in the MPC while achieving a similar level of enhancement to the performance of the controller.

It must be noted that the basis of this method arises from the capability of the controller to maintain the quadrotor near its reference trajectory. Therefore, if a sudden external disturbance perturbs the system away from the reference trajectory, then the pre-computed model corrections are no longer accurate. In these circumstances, the GP predictions can be made from the control trajectory produced in the previous time step, until the quadrotor is close enough to the reference trajectory to utilize the pre-computed model corrections.

## V. Conclusion

In this work, we have presented a computationally efficient data-driven MPC approach for agile quadrotors. The GPs learn the residual dynamics from the collected data points and provide corrections to a simple dynamic model of the system to enhance the tracking performance of the controller. Only the first control node is corrected online using the current measurements. To reduce the computational burden, for the remaining nodes, we leverage on the controller's capability to track the reference trajectory accurately and predict the model corrections in advance using the reference trajectory. Simulation results have demonstrated similar levels of improvements compared to the GP-based controller from the literature where all corrections are computed online. In particular, the performance of the two controllers has been equivalent in a noisy setting, imitating the inaccurate state measurements of the system in a real-time application. This demonstrates a method to implement GP correction models to an MPC framework without imposing a significant computational burden.

As a future study, the improved computational efficiency of the GPs in the MPC framework may allow the utilization of the GPs to be extended to state estimators, *e.g.*, moving horizon estimation, as having a better knowledge of the system's states will ultimately improve the controller's performance.